\documentclass{article}
\usepackage{spconf,amsmath,graphicx}
\usepackage[utf8]{inputenc}

\graphicspath{ {./figures/} }
\usepackage{epsfig} % for postscript graphics files
\usepackage{subfig}
\usepackage{xcolor}
\usepackage{url}
\title{Adaptive Gray World-Based Color Normalization of Thin Blood Film Images}

\name{F. Boray Tek\textsuperscript{1*} \thanks{*performed the work while at University of Westminster} \thanks{Corresponding email: boray.tek@isikun.edu.tr}, Andrew G. Dempster\textsuperscript{2}, İzzet Kale\textsuperscript{3}}
\address{{\textsuperscript{1} Department of Computer Engineering, Işık University, İstanbul, Turkey}\\ \textsuperscript{2}School of Electrical Engineering and Telecommunications, University of New South Wales, Sydney, Australia \\ \textsuperscript{3}	Applied DSP and VLSI Research Group, University of Westminster, London, UK}
\begin{document}
\maketitle

%%%%%%%%%%%%%%%%%%%%%%%%%%%%%%%%%%%%%%%%%%%%%%%%%%%%%%%%%%%%%%%%%%%%%%%%%%%%%%
%% A B S T R A C T
%%%%%%%%%%%%%%%%%%%%%%%%%%%%%%%%%%%%%%%%%%%%%%%%%%%%%%%%%%%%%%%%%%%%%%%%%%%%%%

\begin{abstract}
This paper presents an effective color normalization method for thin blood 
film images of peripheral blood specimens. Thin blood film images can easily 
be separated to foreground (cell) and background (plasma) parts. The 
color of the plasma region is used to estimate and reduce the 
differences arising from different illumination conditions. A second stage 
normalization based on the database-gray world algorithm transforms the 
color of the foreground objects to match a reference color character. The 
quantitative experiments demonstrate the effectiveness of the method and 
its advantages against two other general purpose color correction methods:
simple gray world and Retinex.
\end{abstract}

%%%%%%%%%%%%%%%%%%%%%%%%%%%%%%%%%%%%%%%%%%%%%%%%%%%%%%%%%%%%%%%%%%%%%%%%%%%%%%
%% I N T R O D U C T I O N
%%%%%%%%%%%%%%%%%%%%%%%%%%%%%%%%%%%%%%%%%%%%%%%%%%%%%%%%%%%%%%%%%%%%%%%%%%%%%%

\section{Introduction\label{sec:intro}}
Microscopy examination of blood specimens is still the gold standard for the 
diagnosis of several diseases. However, it is usually a tedious and subjective
task to perform manually. On the other hand, automated diagnosis systems aim to
replicate diagnostic expertise by specifically tailored image processing, 
analysis, and pattern recognition algorithms \cite{Tek2010}. 
However, image variations that are commonly neglected by humans generate
problems for automated algorithms.

An image acquired from a stained blood specimen (thick or thin) using a 
conventional light microscope can have several conditions which may affect 
the observed colors of the cells, plasma (background), and stained objects. 
These conditions may be due to the microscope components such as: 
different color characteristics of light sources, intensity adjustments, 
or color filters; use of different cameras or 
different settings in the same camera such as exposure, aperture diagram, 
or white balance settings. It is possible to reduce these effects by 
calibration. However, variations can also be caused in slide preparation such as
use of different stain concentrations, exposure to different 
staining durations, or non-uniform staining.

There are various studies addressing the calibration and color constancy 
issues for general imaging \cite{KobusBarn99}. However,
the field literature for microscope imaging is quite limited. 
The general purpose methods of color correction is not appropriate for 
microscope imaging because first the Lambertian surface model does not fit to 
microscope imaging, 2) the use of the reference color 
charts are not practical. Simply, because the sensor (or human eye) does not receive 
the light reflecting from a surface; it is the attenuated light that is left 
from the object's (i.e. specimen's) absorption. In fact, image formation 
of the stained slides with light microscopes can be appropriately modeled
with the ``Beer-Lambert Law'', which states that there is a linear relationship
between the concentration, thickness of illuminated media, and the ``absorbance''
\cite{lee_2005}. In addition, the reference color patches 
(as proposed for other medical imaging applications, e.g.\ \cite{grana_05},
is not practical for microscopes. Moreover, there is still the human factor in preparation of the 
slides which results in non-standard and inhomogeneous staining 
concentrations and colors \cite{yagi_05}. 

The problem of non-standard preparation of the slides (specimen) was 
addressed in \cite{abe_04}. To correct under/\-over staining conditions of 
the slide, Abe et al.\ obtained the spectral transmittance by a multi-
spectral camera, and mathematically modeled the 
relation between the transmittance and the amount of stain (dye) for each 
pixel using the Beer-Lambert Law and Wiener inverse estimation.
However, the method required a multi-spectral camera to operate, and
variations caused by different light sources were not addressed.

In \cite{boray_siu_06} we proposed a practical method which 
exploits the special characteristics of the peripheral thin blood film images 
that are easily separable to foreground and background regions. However, the 
method was presented briefly as a procedure and no quantifiable evaluation was given.
In this study, we reformulate this color normalization algorithm, and 
compare it to two other general purpose color correction algorithms:
simple gray world \cite{KobusBarn99} and Retinex \cite{funt_00}.
 
\section{Color Constancy}\label{chap5:sec.method}
The primary aim of color constancy is to obtain an illumination and imaging sensor
independent color representation of scenes. In general, 
these two factors are studied separately: illumination change and camera calibration.
Sometimes the latter is not considered (as in this study) 
when the imaging sensor or camera is unknown or its calibration is 
not possible.

There are some different models of illumination change in the literature \cite{KobusBarn99}.
The ``diagonal model'' is a simple and satisfactory model 
which assumes that there is a diagonal $3\times3$ linear transformation matrix 
($\mathbf{M}$) which maps the RGB response of 
an unknown illuminant $\textbf{p}^{u}= (r^{u},g^{u},b^{u})$ to the RGB 
response of a canonical known illuminant $\textbf{p}^{c}= (r^{c},g^{c},b^{c
})$: ($\mathbf{p}^{c}=\mathbf{M}\mathbf{p}^u$). If the transformation matrix
is assumed to be diagonal (\ref{equ.1}), its non-zero elements ($m_{ii}$) 
can be calculated by simple scaling $p^{c}_{i}/p^{u}_{i}$ where $i \in\{r,g,b\}$.

\begin{equation}\label{equ.1}
\mathbf{M} =
\left[ \begin{array}{ccc}
	m_{rr} & 0 &0  \\
 0& m_{gg} & 0 \\
0& 0 & m_{bb}
\end{array} \right] 
\end{equation} 
where the non-zero elements of the diagonal matrix $m_{rr}$, $m_{gg}$, and $m_
{bb}$ are the illumination factors. If the illumination is assumed to be 
uniform, using the diagonal model, an image of unknown illumination $\mathbf{I
}^u$ with RGB color vector can be simply transformed to a known illuminant 
space $\mathbf{I}^c$ by multiplying all the pixel values with the diagonal 
matrix $(\mathbf{I}^c = \mathbf{M}\:\mathbf{I}^{u})$.

The simplest color constancy algorithm is based on the ``gray world'' assumption
that the average value of the scene is stable (i.e.\ gray) under the same 
illumination. Therefore, a deviation of the average value of the image is expected
to show the illumination change. The different interpretations of this assumption can 
lead to different approaches. For example, the average value of the scene can 
be assumed to be the level that is a portion (e.g.\ half) of the maximum 
possible intensity value of each channel. Thus, using the diagonal model, the 
illuminant factors can be calculated as in (\ref{equ.2}):
\begin{equation}\label{equ.2}
m_i= \frac{G_i}{\mu^u_i}  \qquad (\mu^u_i=\frac{1}{N}\sum_{N}I^u_i) \qquad  i=
\{r,g,b\} 
\end{equation}
where $G_i$ is the constant (assumed) gray value and the $\mu^u_i$ is the 
mean of the channel $I^u_i$ which has $N$ total pixels. Alternatively, the 
\emph{gray value(s)} can be defined with respect to the recorded values of 
the scene under a known illuminant. This is called \emph{the database gray-
world algorithm} \cite{hordley_04,Barnard_02} and can be represented as in 
(\ref{equ.3}):
\begin{equation}\label{equ.3}
m_i= \frac{\mu^c_i}{\mu^u_i} \qquad   i=\{r,g,b\}
\end{equation}
Thus the illumination factors are calculated by the ratios of the average 
values of the each channel of the reference ($\mu^c_i$) to those of unknown 
($\mu^u_i$). Hence the term \emph{database} refers to calculation of the 
\emph{gray value} from a reference set of images.

For ordinary images, the illumination estimation or color normalization 
based on the gray world assumption with either of the calculations yields 
poor results compared with more sophisticated algorithms such as gamut 
mapping, Retinex or color by correlation \cite{KobusBarn99,hordley_04,Barnard_02}.
Obviously, the weakness of the gray world algorithm arises from the 
fact that it is hard to assume that unconstrained physical 
scenes to average to a universal gray value.

\subsection{Proposed Methodology}

The images subject to this study are simpler to generalize because
they are composed of two basic parts: plasma and 
the rest which contains mostly red blood cells. Moreover, these parts 
can be separated by thresholding. Plasma or 
more appropriately background of the peripheral thin blood film image can 
be assumed to be a colorless transparent region which reflects the 
chromaticity of the absorbed light. Therefore, by calculating 
the average pixel values of the different color channels in the plasma (background)
region of the images, we can obtain an estimate for the 
illuminant $RGB$, which can be used to normalize each 
channel (of the whole image) respectively.

This normalization cannot correct the variation due to staining because,
as explained before, illuminant and staining are two independent sources of variation.
However, now the illuminant variation is saturated, we can assume the perceived
difference in foreground objects is due to staining concentration.
Thus, it is possible to transform foreground pixels using Eq.\ref{equ.3}, 
where the gray values under known illuminant can be used as reference or 
\emph{database} gray values.

Therefore, initially, input images must be separated into foreground and 
background regions. The separation (i.e.\ binarization) can be performed 
using Otsu's thresholding or the method proposed by \cite{rao_03}, which provides
advantages. This method uses area morphology 
to estimate size of the cells and then extracts foreground objects and 
estimates individual histograms for the foreground and background regions. 
These histograms are used to obtain two separate thresholds to perform a 
morphological double thresholding operation \cite{Soille_book}. 
 
After separating the input ($\mathbf{I}^u_{i}$) channels ($i \in \{r,g,b\}$), 
foreground $(\mathbf{I_f}^{u}_{i})$ and background $(\mathbf{I_b}^{u}_{i})$ 
images are obtained. Then the proposed color normalization is performed as 
follows:

\begin{enumerate}
 	\item  Calculate $
\mathbf{M^b}$ using $(\mathbf{I_b}^{u}_{i})$ channel averages:\\
 $m^b_{i} = \frac{255}{\mu^{\mathbf{I_b}^{u}_{i}}}$.
	\item Transform the whole image: $\mathbf{I^1}$ = $\mathbf{M^b}$ $\mathbf{I^
u}$ using Eq.\ref{equ.2}.
	\item Calculate $\mathbf{M^f}:$ $m^f_{i}$ using Eq.~\ref{equ.3}
 with $(\mathbf{I_f}^{1}_{i})$ and the reference image foreground channels $
\mathbf{I_f}^{c}_{i}$. 
	\item Transform only the foreground channels: $\mathbf{I_f}^{2} = \mathbf{M^
f}\mathbf{I_f}^{1}$
	\item Replace the foreground channels of $\mathbf{I^1}$ with $\mathbf{I_f}^{
2}$ to obtain the final color normalized output image $\mathbf{I^2}$. 
\end{enumerate}

\begin{figure}[tbh]
\centering
\includegraphics[width=\columnwidth]{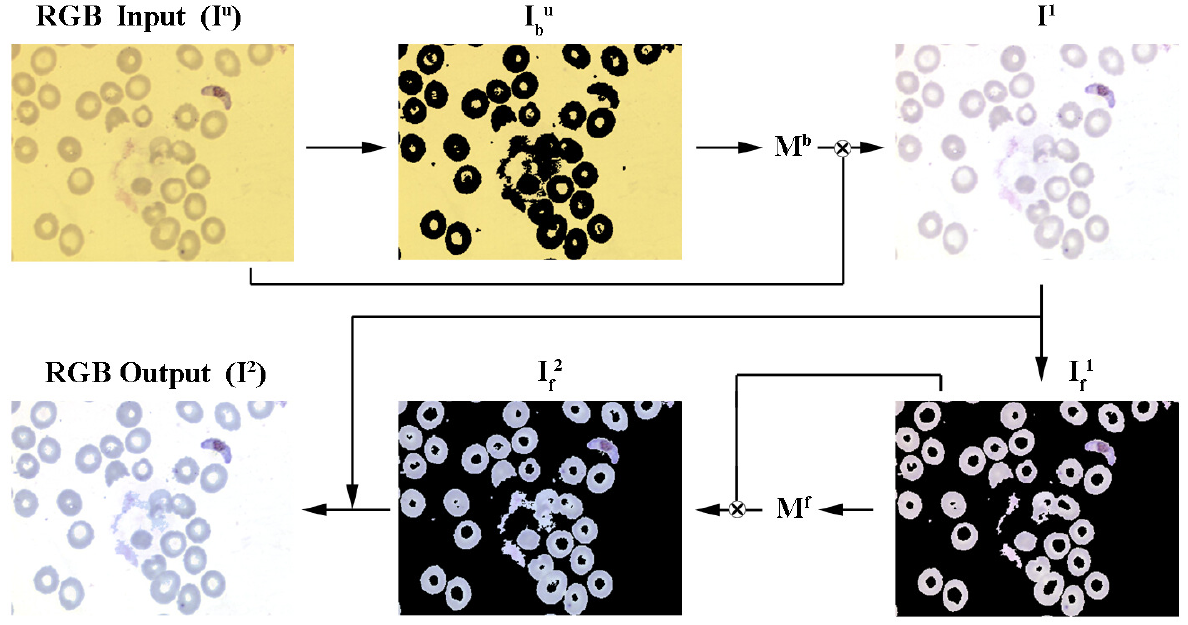}
\caption[Color normalization steps]{Color normalization steps: 1) Input 
RGB image, 2) background pixels are separated, 3) input image is normalized 
by its background average value (for each channel $RGB$) ($I^1$), 4) 
foreground pixels are extracted $({I_f}^{1}_{r, g, b})$, 5) the foreground 
pixels are normalized $\mathbf{I_f}^{2}$, 6) the normalized foreground pixels 
of each channel are replaced with the ones in $I^1$ to obtain output (color 
corrected image) $I^2$.}
\label{fig.col_norm_flow}
\end{figure}

The procedure is demonstrated with an example image in Fig \ref{fig.col_norm_flow}.
Following the image binarization using the area morphology 
technique \cite{rao_03}, the background pixels are extracted. Using the 
background pixel averages in (\ref{equ.3}), input image is normalized 
with respect to the background average color values (estimated illumination 
color). Then from this image only the foreground pixels are extracted and 
normalized according to the $R,G,B$ values that were determined by a reference set.
The final result is obtained by replacing the normalized foreground pixels. 

%\ref{As a result of the process, the output histograms are shifted and 
%aligned with the reference set histograms.}

%\ref {Put here the chart.} 

\section{Experimental Results}
In order to evaluate the proposed method we have collected 15 images of the 
same thin blood film (standard Giemsa-stained) field using 
a Brunel Sp 200 (Brunel, UK) microscope and a Canon A60 digital camera 
coupled with Unilink (Brunel, UK) adapter. 
In order to construct the database reference (gray) values, a reference set of 12 
thin color film images (of different fields) was prepared; however, the 
images were not chosen in terms of quality. The only criterion was that the 
images have to be acquired with the same illumination and camera settings. 
The illuminant or camera settings were not recorded or required. After an 
initial normalization with respect to the background channel average values
$R,G,B$, the foreground $R,G,B$ channel average values were calculated for 
each channel ($\mu^c_r = 183, \mu^c_g=189, \mu^c_b=214$), respectively,  
which were then used in the normalizations of the foreground images. 
To provide a comparison we implemented simple gray world normalization 
algorithm and also used a MATLAB$^{TM}$ implementation of Retinex \cite{funt_00}.
However, in order to work effectively Retinex implementation \cite{funt_00}
requires log-quantized input images, which was not possible to provide 
with Canon A60. However, we applied pre-normalization with respect to the 
minimum and maximum channel values. The number of iterations parameter for 
Retinex was set to 4. To measure and compare the performance, 
we followed the related works \cite{hordley_04,Barnard_02}
and calculated the RMS (root mean square) of the angular difference:

\[
	E_\theta = acos(\frac{\mathbf{p_1} . \mathbf{p_2}}{\left\|\mathbf{p_1}\right\|\left\|\mathbf{p_2}\right\|})
\]
where $p_1$ and $p_2$ are two different output pixel $RGB$ triples.

\begin{figure}[tbh]
	\centering
\includegraphics[width=\columnwidth, clip  ]{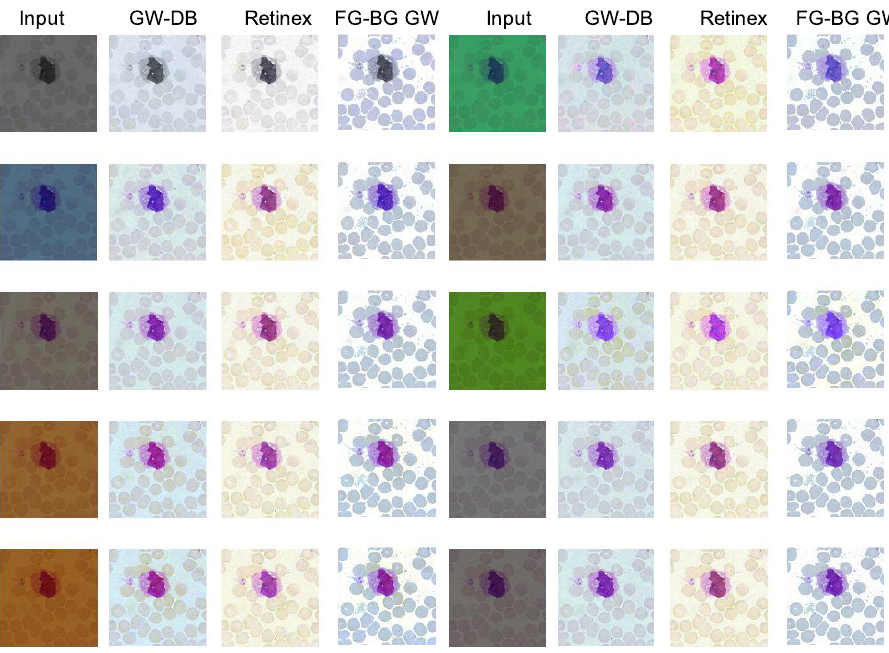}
	\caption{Different input images of the same field under different imaging 
conditions and output of gray world (GW), database gray world (GW-DB), 
Retinex, and proposed gray world algorithm (FG-BG GW). }
\label{fig.comp}
\end{figure}

\begin{figure}[tbh]
	\centering
\subfloat[]{\includegraphics[width=0.75\columnwidth, clip  ]{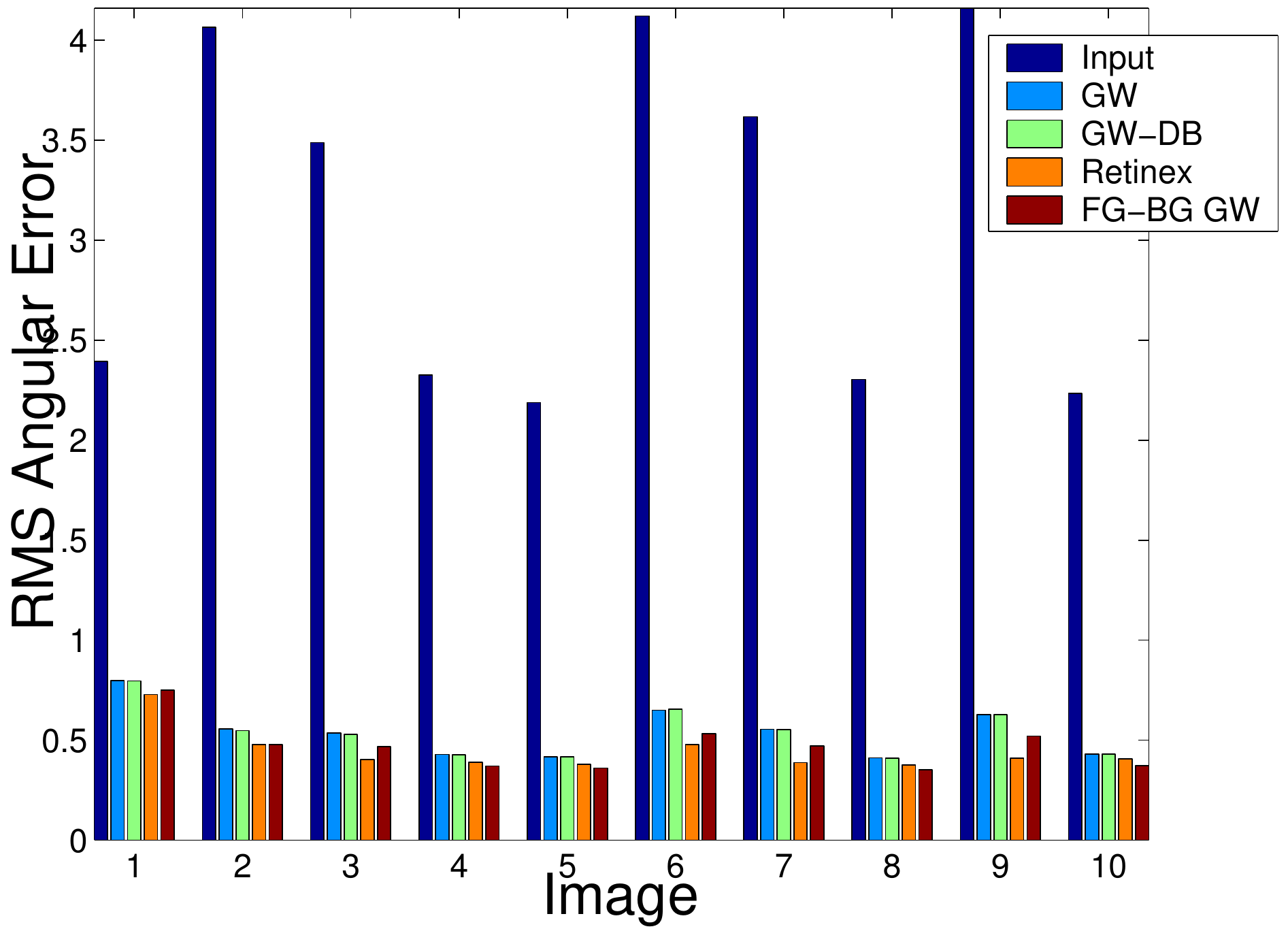}\label{chap5:fig.6a}}

\subfloat[]{\includegraphics[width=0.75\columnwidth, clip  ]{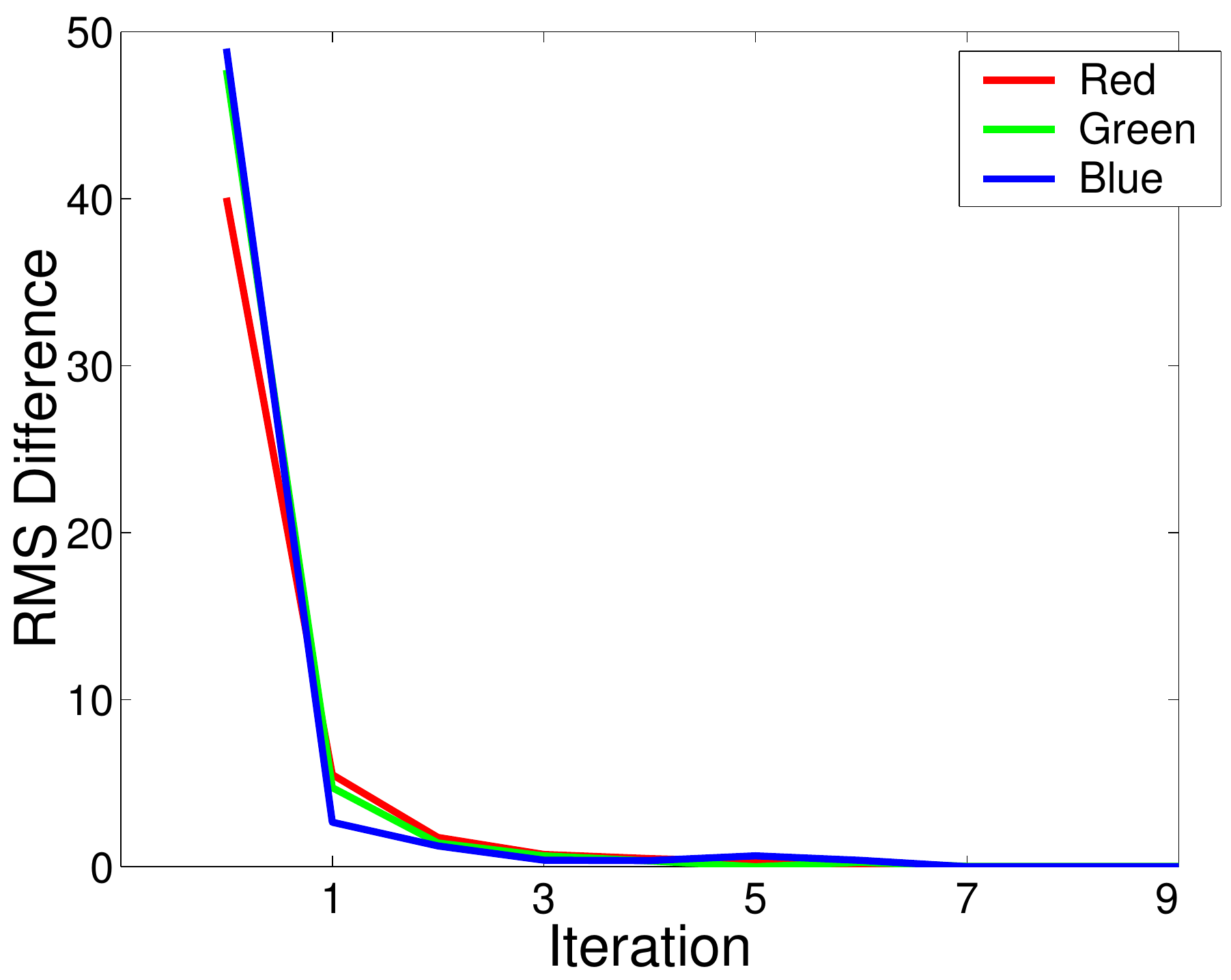}\label{chap5:fig.6b}}
	\caption{(a) Sum of RMS angular differences between color corrected outputs 
of different algorithms, (b) RMS difference versus iteration of the proposed
method for an input and output image.}
	\label{chap5:fig.6}
\end{figure}

Fig. \ref{fig.comp} shows input images and corresponding outputs from the 
three different algorithms: database gray world (GW-DB), Retinex, and 
the proposed (foreground-background separation based) gray world (FG-BG GW). 
The input images contain significant color casts which result from the different 
illumination and sensor color balance. In general all three algorithms can remove the color casts
in majority of cases, whereas our algorithm can recover colors even in the desaturated
image. Figure \ref{chap5:fig.6a} visualizes
the sum of RMS angular differences for each input images, 
where the RMS difference values for the unprocessed inputs are provided
as a reference. It can be seen that the both of Retinex and FG-BG GW algorithms
performed best in five images whereas GW-DB, whereas GW performed poor in most of the 
cases. 

Note that, the presented algorithm is not proposed as an iterative process, 
however it is expected to be idempotent and make no difference if applied to an 
input image with correct colors. Figure \ref{chap5:fig.6b} shows the RMS 
difference between an input-output pair of the FG-BG GW algorithm 
with respect to consecutive applications. 
It can be observed the significant difference is obtained in 
the first run, whereas the algorithm converges rapidly, as the ratios (i.e.\ diagonal 
entries of $\mathbf{M}$) approach to $1$. Moreover, we have tested our algorithm
on arbitrary thin blood film images of different blood abnormalities 
from arbitrary internet sources and observed that it provides robust color correction.

%\begin{figure}
	%\centering
%\subfigure[]{\includegraphics[scale =0.16, clip  ]{col_web2_in.eps}\label{chap5:fig.9a}}
%\subfigure[]{\includegraphics[scale =0.16, clip  ]{col_web2_out.eps}\label{chap5:fig.9b}}
%\subfigure[]{\includegraphics[scale =0.16, clip  ]{col_web3_in.eps}\label{chap5:fig.9c}}
%\subfigure[]{\includegraphics[scale =0.16, clip  ]{col_web3_out.eps}\label{chap5:fig.9d}}
%\subfigure[]{\includegraphics[scale =0.16, clip  ]{col_web4_in.eps}\label{chap5:fig.9e}}
%\subfigure[]{\includegraphics[scale =0.16, clip  ]{col_web4_out.eps}\label{chap5:fig.9f}}
%\subfigure[]{\includegraphics[scale =0.16, clip  ]{col_sickle_in.eps}\label{chap5:fig.7c}}
%\subfigure[]{\includegraphics[scale =0.16, clip  ]{col_sickle_out.eps}\label{chap5:fig.7d}}
%\subfigure[]{\includegraphics[scale =0.16, clip  ]{col_6a_in.eps}\label{fig.7a}}
%\subfigure[]{\includegraphics[scale =0.16, clip  ]{col_6a_out.eps}\label{fig.7b}}
%\subfigure[]{\includegraphics[scale =0.16, clip  ]{col_4a1_in.eps}\label{fig.7c}}
%\subfigure[]{\includegraphics[scale =0.16, clip  ]{col_4a1_out.eps}\label{fig.7d}}
	%\caption{color normalization - example images from external sources: 
%Images from Google search engine, input-output pairs show (a)-(b): Wuchereria 
%bancrofti, (Filariasis parasite) (c)-(d) P. Falciparum (e)-(f) Trypanosoma 
%cruzi (i.e.\ Chagas Disease), (g)-(h) sickle cell, (i)-(j) P. Falciparum, (k)-
%(l) P. Vivax.}	
%\label{chap5:fig.9}
%\end{figure}

\section{Conclusions}
We presented a novel color correction method for peripheral thin blood film 
images based on the gray world and database gray world assumptions. 
Our method assumes that plasma (background) is observable in the input image. It further 
assumes the background is colorless under ideal microscope lighting conditions; it 
scales the whole image to shift the plasma color towards maximum possible 
level to reduce the effects of illumination. The foreground color correction 
assumes that the foreground scene in the image is distributed normally around 
an average value, which is then scaled to match a value that is calculated 
from a reference image set of desired color.

We show that it is better than the simple gray world algorithms,
whereas it has similar or better performance to Retinex
(in terms of RMS angular error). A color correction algorithm must be idempotent 
if applied to an already normalized image. Our algorithm 
converges rapidly to a stable state after the first application.

In addition, our algorithm is favorable to Retinex, because the latter requires
iterations which are computationally more demanding. However, it is possible
to investigate whether the proposed method can be improved with use of Retinex 
instead of the simple gray world assumption. Moreover, it would be useful to test and compare the performance of the methods in terms of their performance as a preprocessing step to an automatic analysis algorithm.

%\vfill
%\pagebreak

\bibliographystyle{IEEEbib-short}
\bibliography{PhD_v5}
\vfill

\end{document}